\documentclass[letterpaper]{article}
\usepackage{iccc}
\usepackage{times}
\usepackage{helvet}
\usepackage{booktabs}
\usepackage{multirow}
\usepackage{array}  
\usepackage{tabularx}
\usepackage{url}
\usepackage{amsmath}
\usepackage{makecell}
\usepackage{graphicx}
\usepackage{float}
\usepackage{xcolor}
\usepackage{courier}
\usepackage{url}
\expandafter\def\expandafter\UrlBreaks\expandafter{\UrlBreaks
  \do\a\do\b\do\c\do\d\do\e\do\f\do\g\do\h\do\i\do\j%
  \do\k\do\l\do\m\do\n\do\o\do\p\do\q\do\r\do\s\do\t%
  \do\u\do\v\do\w\do\x\do\y\do\z\do\A\do\B\do\C\do\D%
  \do\E\do\F\do\G\do\H\do\I\do\J\do\K\do\L\do\M\do\N%
  \do\O\do\P\do\Q\do\R\do\S\do\T\do\U\do\V\do\W\do\X%
  \do\Y\do\Z}
\title{Abductive Computational Systems: Creative Abduction and Future Directions}
\author{Abhinav Sood\\
University of Sydney\\
Sydney, Australia\\
abhinav.sood@sydney.edu.au\\
\And
Kazjon Grace \\
University of Sydney\\
Sydney, Australia\\
kazjon.grace@sydney.edu.au\\
\And
Stephen Wan\\
Data61, CSIRO\\
Sydney, Australia\\
stephen.wan@data61.csiro.au\\
\And
Cecile Paris \\
Data61, CSIRO\\
Sydney, Australia\\
cecile.paris@data61.csiro.au\\
}
\begin{document} 
\maketitle
\begin{abstract}
\begin{quote}
Abductive reasoning, reasoning for inferring explanations for observations, is often mentioned in scientific, design-related and artistic contexts, but its understanding varies across these domains. This paper reviews how abductive reasoning is discussed in epistemology, science and design, and then analyses how various computational systems use abductive reasoning. Our analysis shows that neither theoretical accounts nor computational implementations of abductive reasoning adequately address generating creative hypotheses. Theoretical frameworks do not provide a straightforward model for generating creative abductive hypotheses, and computational systems largely implement syllogistic forms of abductive reasoning. We break down abductive computational systems into components and conclude by identifying specific directions for future research that could advance the state of creative abductive reasoning in computational systems.
\end{quote}
\end{abstract}

\section{Introduction}

Abductive reasoning is a form of logical reasoning that deals with explanations for hypotheses, conjectures, anomalies, etc. For example, consider a person who returns to their house to find their belongings scattered around, along with a broken window. If they had left the house clean, they could abduce that a burglar might have broken into their home while they were out. Charles S. Peirce first formulated this form of reasoning as \cite{peirce1958collected}:
\begin{enumerate}
    \item A surprising fact O is observed.
    \item But if [hypothesis] H were true, O would follow.
    \item Thus, there is a reason to suspect that H is True.
\end{enumerate}

Unlike deductive reasoning, abductive reasoning does not aim to preserve the truth of the derived propositions. Instead, pragmatically, abductive reasoning seeks to explain surprising observations where other forms of reasoning might not be possible. In a computational creativity context, this contrasts with how surprise typically provides an evaluation of creativity \cite{grace2014expect}. Instead, abduction posits that surprises are an initiating event: they are the unexplainable observations for which abduction provides an explanation. This ability to generate plausible explanations in the face of surprise makes abductive reasoning particularly valuable in both scientific, design-related, and artistic contexts: scientists rely on it to formulate hypotheses that might explain unexpected phenomena, while designers use it to envision solutions that address observed needs without complete information about the problem space. Similarly, artists employ abductive reasoning when interpreting ambiguous visual elements and unexpected results, using these surprises as catalysts for developing artistic expressions.

Thus, understanding how computational systems use abductive reasoning can help us identify properties that may increase their utility. To the best of our knowledge, no existing review examines how abductive reasoning is implemented across computational systems in these diverse fields. The main contributions of this paper are:
\begin{enumerate}
    \item We review and connect literature on abductive reasoning that exists across different, often unlinked fields.
    \item We analyse how computational systems implement abductive reasoning and propose a breakdown that identifies simple components for advancing creative abduction.
\end{enumerate}

\section{Abductive Reasoning in Theory}
\subsection{Epistemological Background}
Since Pierce's formulation of abduction, several philosophers have further developed the concept of abductive reasoning. \citeauthor{gabbay2005reach} (\citeyear{gabbay2005reach}) discuss the ignorance-preserving nature of abduction. Ignorance-preserving means that abductive reasoning does not generate a hypothesis that completely resolves the surprising observation. Instead, uncertainty remains about whether the hypothesis is true even after a hypothesis is produced. So, what use does the hypothesis provide? It is a reasoned basis for future action \cite{gabbayAdviceAbductiveLogic2006}.

By abductive reasoning, we are not referring to inference to the best explanation (IBE). While abduction is just the reasoning process to create and select a possible hypothesis, IBE is an iterative process that might involve deductive, abductive and inductive reasoning to select and arrive at the best hypothesis available. A model for IBE in diagnostic reasoning is presented in \cite[23]{magnaniAbductionReasonScience2001}. This model uses abduction to propose diagnostic hypotheses based on clinical evidence. Observed clinical data corroborate hypotheses through induction, and the deductive implications of the hypothesis can provide us with expectations of what our data should look like. These expectations form the basis for future surprises that trigger further abduction.

There are several formalisations of abductive reasoning. The Gabbay Woods Schema (GW-Schema) \cite[pg. 198]{gabbayAdviceAbductiveLogic2006} aims to capture several properties of Peircean abduction. Although the GW-Schema illustrates the underlying syllogistic logic of abductive reasoning, it suffers from two major problems. Firstly, it does not explain how the hypotheses themselves are generated. Second, it does not concretely define the conditions that a hypothesis needs to meet to be considered worthy of investigation. Some of the conditions often discussed in the literature \cite{sep-abduction} are simplicity, coherence and explanatory power. However, no consensus exists on whether these specific conditions are necessary. For example, in \cite[pg. 139]{Woods2017}, Woods rejects simplicity and "sees no reason why truth favours the uncomplicated". The issues of coming up with a hypothesis and of selecting a hypothesis among candidates based on specific conditions are referred to as the fill-up and cut-down problems \cite[pg. 6]{magnaniAbductiveStructureScientific2017} respectively. Magnani discusses how the solutions to these problems are "stunningly contextual" \cite[pg. 10]{magnaniAbductiveStructureScientific2017}. While his proposed eco-cognitive model provides us with an understanding of abductive reasoning in more creative, scientific contexts, it does not provide a computable formal structure.

When looking for computable formal structures, some might contest that Abductive Logic Programming (ALP) \cite{kakas1992abductive} can fill the gap. ALP extends standard logic programming by introducing abducible predicates. These frameworks allow systems to derive explanations by backward-chaining from observations to possible causes, constrained by integrity constraints that filter out implausible explanations. ALP has been shown to be effective for diagnostic reasoning \cite{koitz2018applying}, but typically operates in closed, well-defined domains and struggles with more open-ended abduction. 

The differing theoretical approaches to abductive reasoning have led Woods to argue that "The foundational work for a comprehensive account of abductive reasoning still awaits completion"  \cite[pg. 138]{Woods2017}. Unfortunately, this means that the epistemological literature on abductive reasoning does not sufficiently address how you can functionally perform the more complex forms of abductive reasoning. As such, it might be worthwhile to look at how creative abductive reasoning emerges in two domains where creativity and hypotheses are central: science and design.

\subsection{Creative Abductive Reasoning in Science and Design}
In the philosophy of science, abductive reasoning is traditionally associated with creative processes in the context of discovery \cite{sep-scientific-discovery}. Here, the "context of discovery" refers to the initial generation of theories in science through the conception of hypotheses and research ideas. Logically abductive reasoning deals with producing explanations for surprising observations. We propose that this process of producing explanations is relevant throughout the scientific processes at two distinct levels.

In the "context of discovery", it is relevant macroscopically; abductive reasoning serves as a guiding framework for the entire scientific process, shaping what hypotheses and ideas are investigated. Microscopically, abductive reasoning is used within individual experimental stages of the scientific process. To illustrate this, consider the documentation of the discovery of the urea cycle in \cite{kulkarni1988processes}. Macroscopically, after observing the unexpected effect of ornithine in enhancing urea production, Krebs abductively reasoned that ornithine might play a central role in a cyclical process of urea formation. This explanation accounted for the surprising observation and led him to restructure his entire research program around understanding ornithine's role in urea synthesis. Note that he "did not have a clear hypothesis of a mechanism to account for it [ornithine effect]" that he was testing \cite{kulkarni1988processes}. Instead, a rougher hypothesis was sufficient to guide his investigation. Microscopically, Krebs employed abductive reasoning at specific experimental junctures. When he discovered that the effect was unique to ornithine (as chemical derivatives failed to produce similar results), he generated the explanatory hypothesis that ornithine possessed specific structural properties essential to urea formation.

In design, abduction takes on different characteristics when compared to abductive reasoning in science. Design theorists have identified abduction as the fundamental reasoning pattern for moving from function to form, from “what is needed” to “how to do it” \cite{kroll2015abduction}. \citeauthor{roozenburg1993pattern} (\citeyear{roozenburg1993pattern}) made a distinction between explanatory abduction (explaining existing phenomena) and innovative abduction (creating new solutions), and argued that design is primarily the latter. This innovative form of abduction in design is different from scientific abduction as it aims to create entirely new artefacts to fulfil desired functions. \citeauthor{koskela2018role} (\citeyear{koskela2018role}) further elaborated on this view, saying that abduction in design is not just an inference type but a property of many inferences throughout the design process. Using a SAPPhIRE‐based model of abductive reasoning in design, \citeauthor{bhatt2021analyzing} (\citeyear{bhatt2021analyzing}) provide a framework to explain how designers use abductive reasoning to bridge the gap between desired functions and proposed solutions. These models are useful to understand the nature of abduction, but can they be automated? To further understand how abductive reasoning takes place practically, we look at how computational systems implement abductive reasoning. Just as \cite[pg. 17]{boden2004creative} argues that computational ideas can inform our understanding of human creativity; computational ideas can inform our understanding of abductive reasoning. 

\section{Abductive Computational Systems}

\subsection{Natural Language Processing (NLP)}
For modern computational systems, recent investigations of abductive reasoning have largely focused on introducing it in Large Language Models (LLMs). Large Language Models (LLMs) are neural networks that generally utilise the transformer architecture \cite{transformers} to model natural language probabilistically. One of the first papers to investigate abductive reasoning with LLMs was \cite{Bhagavatula2020Abductive}. They considered how well LLMs perform abductive reasoning in everyday situations derived from datasets of short stories. The authors created two tasks to test how well LLMs performed. Alpha NLI tested if LLMs could select the correct abductive hypothesis (given a choice of 2) that fits observations that occur before and after the hypothesis. The language model they trained had an accuracy of 68.9\%, with humans identifying the correct hypothesis in 91.4\% of cases. In our experiments, we found that LLaMa 3.1 70B-Instruct \footnote{{\color{blue}\text{https://huggingface.co/meta-llama/Llama-3.1-70B-Instruct}}}  was able to achieve an accuracy of 86.2\% with a zero-shot prompt on the test set of the Alpha NLI task, indicating that modern LLMs might perform well in identifying correct abductive hypotheses in everyday situations. These results are merely indicative, as there might be data leakage from the test set, but they point towards the possibility of effective abductive reasoning through LLMs. Beyond everyday situations, \citeauthor{zhao-etal-2024-uncommonsense} (\citeyear{zhao-etal-2024-uncommonsense}) train LLMs to generate uncommon explanations in everyday situations.  \citeauthor{tian2024macgyver} (\citeyear{tian2024macgyver}) constructs MacGyver-like creative problems to train LLMs. These studies focus on extending the abductive capabilities of LLMs by creating datasets for specific tasks. Such datasets that exclusively focus on abductive reasoning are not yet prevalent in science and design. 

Many abductive systems in NLP also focus on knowledge bases. AbductionRules \cite{young-etal-2022-abductionrules} tests how well transformers reason over logical knowledge bases. \citeauthor{bai-etal-2024-advancing} (\citeyear{bai-etal-2024-advancing}) use knowledge graphs to generate hypothesis-observation pairs. 
There is also research on hypothesis generation in NLP that does not explicitly mention abductive reasoning. These systems have their own host of issues.

In the case of \cite{si2025can}, several limitations of the AI-generated research ideas (inappropriate baselines, unrealistic assumptions, etc.) were pointed out, which brings into question whether the novel ideas generated are usable in practice. With the AI Scientist \cite{lu2024aiscientistfullyautomated}, LLMs themselves evaluated the quality of AI-scientist-generated research papers, and such modes of evaluation have been called into question \cite{koo2024benchmarking}.

Beyond commonsense reasoning, abductive reasoning has also been used in several computationally creative systems. 
\begin{table}[!h]
\label{tab:systematic_review}
\renewcommand{\arraystretch}{1.2}
\begin{tabular}{m{0.3cm}p{1.8cm}p{5cm}}
\hline
\textbf{\#} & \textbf{System} & \textbf{Abductive Approach} \\
\hline
\multirow{3}{*}{(1)}  & The Painting Fool \cite{colton2010painting} & Identifies constraints among partial rectangles, extends patterns to produce abduced versions \\
\hline
\multirow{3}{*}{(2)}   & Synthetic Audience \cite{o2011simulating} & Extracts information from incomplete stories to build a mental model, using abduction to infer character goals \\
\hline
\multirow{3}{*}{(3)}   & MILA-S \cite{goel2015impact} & 
Prompts students to engage with abductive explanations through ecological model building. \\
\hline
\multirow{3}{*}{(4)}   & ANGELINA \cite{cook2018redesigning} & Uses abductive reasoning with answer set programming to generate game rulesets \\
\hline
\multirow{3}{*}{(5)}   &  Hello \cite{veiga2020hello} & Encourages audience to build narratives through abductive reasoning as the artwork is "incomplete" \\
\hline
\multirow{3.5}{*}{(6)}  & Aris \cite{o2022novelty} & Creates abduced hypothesised formal specifications using homomorphic matching and topological similarity on knowledge graphs \\
\hline
\end{tabular}
\small
\caption{Computational Systems at ICCC that mention Abductive Reasoning}
\end{table}
\subsection{Abductive Reasoning in CC Systems}
To find computational creativity (CC) systems where abductive reasoning was used, we conducted a review of CC conferences. The start (2010) and end (2024) years for our review mark the first and the most recent ICCC conference. To identify systems that used abductive reasoning, we used the search terms "abductive", "abduce", and "abduction". Among the ICCC proceedings, 14 papers contained the search terms. Of these 8 mentioned abductive for purposes of literature review, argumentation or used the term "abduction" in the physical rather than logical sense (mostly kidnappings in narrative generation papers). We were left with 6 CC systems that either used abductive reasoning computationally or were used to elicit abductive reasoning in the users of the system. These systems are discussed in Table 1. 
 
 We also reviewed the proceedings of ACM Creativity and Cognition (C\&C) and Design Computing and Cognition (DCC) conferences for the same search terms. In our review of C\&C and DCC proceedings, we found no computational system implementations: the sole C\&C paper used abduction to identify richness (potential to inspire future ideas), inspirational ideas via keywords, while the 11 DCC papers proposed focused on modelling abduction design and were not specifically implemented as CC systems and centered on design theory. We note that our review is not exhaustive. Many computational systems effectively address abductive problems without explicitly framing their approach as abductive reasoning. While these systems may offer valuable insights, their inclusion is beyond the scope of this short paper.

\subsection{Components of Abductive Computational Systems}
\begin{figure}[h]
    \centering
    \includegraphics[width=\columnwidth]{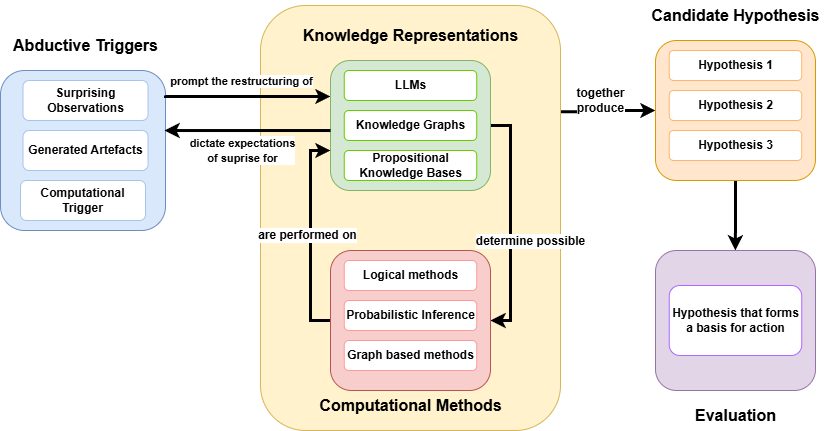}
    \caption{Proposed breakdown for a computational abductive system}
    \label{fig:abductive_framework}
\end{figure}

Through our review of abductive computational systems, we identify four key components that characterise these systems as visualised in Figure \ref{fig:abductive_framework}.   

\subsubsection{Abductive Triggers}:\hspace{-8pt} These act as the starting point for performing abduction. Systems (3) and (5) involve using generated artefacts and constructed models, respectively, to encourage abductive reasoning. The other systems do not necessarily possess abductive triggers in a Peircean sense. Instead, they rely on a computational trigger to perform abductive reasoning. For example, (1) conducts abduction when a user is done inputting a partial configuration of rectangles.

\subsubsection{Knowledge Representations}:\hspace{-8pt}
 In the CC abductive systems we covered, knowledge was encoded as knowledge graphs in (6). In (1), knowledge was construed as constraints that eventually form a constraint satisfaction problem. Abductive reasoning performed in NLP systems we introduced earlier also used knowledge representations. \citeauthor{young-etal-2022-abductionrules}(\citeyear{young-etal-2022-abductionrules}) stored knowledge in a propositional form. \citeauthor{bai-etal-2024-advancing} (\citeyear{bai-etal-2024-advancing}) used knowledge graphs. In \cite{Bhagavatula2020Abductive}, since LLMs were generating the resulting abduced hypothesis, LLMs functioned as black-box stores of knowledge. 
 
\subsubsection{Computational Methods}:\hspace{-8pt} Computational methods perform abductive reasoning in computational systems. These methods heavily depend on the knowledge representation in the CC system. (1) uses extensions of constraints. (4) uses answer set programming to perform abduction. (6) performs abductive reasoning through graph-based methods. The LLM-based systems perform abduction through probabilistic inference.

\subsubsection{Evaluating Hypotheses}\hspace{-8pt}:
In the CC systems, the selection process was very specific to the systems. (1) allows users to remove abduced rectangles individually if they are perceived by the user as incorrectly abduced. (6) uses deductive reasoning to verify abduced hypotheses.  These steps fall under hypothesis verification, which is a part of IBE but not a part of our definition of abductive reasoning. In (3) and (5), the human performs the evaluation. Few systems actually use abductive properties to evaluate hypotheses. \citeauthor{dalal-etal-2024-inference} (\citeyear{dalal-etal-2024-inference})  used properties like consistency, depth, drift, coherence and linguistic uncertainty to evaluate hypotheses, but the hypotheses are restricted to syllogistic if-then structures.

\section{Future Directions}
Our analysis reveals a significant gap between theoretical understandings of abductive reasoning and existing computational implementations. While epistemological accounts emphasise the creative nature of hypothesis generation, computational systems primarily implement restricted, syllogistic forms of abduction. That is, future work should focus on developing systems that can perform creative abduction. One promising direction for this is to create datasets for scientific and design-related abduction, similar to datasets that have already been constructed for commonsense and uncommonsense abductive reasoning. Current systems lack robust methods for evaluating abduced hypotheses, particularly those in natural language form. Developing computationally tractable metrics for properties like simplicity, coherence, and explanatory power could advance our ability to evaluate and compare creative abductive systems.

Our breakdown of computational abductive systems also presents opportunities for the exploration of collaborative and co-creative systems in science. Consider that a literature review of co-creative systems produced 92 systems \cite{COFI}; not one was in the domain of science. For abductive triggers, human-machine collaboration could involve the system identifying potentially surprising patterns in data that a human might overlook, while humans could guide the system toward observations that they believe are relevant to scientific questions. For knowledge representations, humans could supply contextual domain knowledge that might be missing from the system's formal representations, while the system could visualise its knowledge structures in ways that reveal unexpected connections.

\section{Conclusion}
This paper covered how abductive reasoning is viewed and discussed across epistemology, science, design and computational systems. Our analysis shows that while abductive reasoning is widely held to be key to creative processes, computational implementations of abduction are restricted to syllogistic forms that leave little room for creativity. By breaking down abductive computational systems into abductive triggers, knowledge representations, computational methods and evaluation, we provide a basis to understand and implement more advanced computational creative abduction systems that close the gap between the theoretical understanding and computational implementation of abduction.

\section{Author Contributions}
Author 1 (AS) conducted the review and prepared the manuscript. Author 2 (KG), Author 3 (SW) and Author 4 (CP) supervised the research and provided feedback and direction for the development of this work.

\section{Acknowledgement}
We gratefully acknowledge the Collaborative Intelligence (CINTEL) Future Science Platform for their support towards this work.

\bibliographystyle{iccc}
\bibliography{iccc}

\end{document}